\documentclass[letterpaper, 10 pt, conference, onecolumn]{ieeeconf}  

\IEEEoverridecommandlockouts                              
\title{\LARGE \bf
Target-Aware Interaction-Guided Reinforcement Learning for Black-Box Node Injection Attacks on Graph Neural Networks
}

\author{Yi Lan$^{1}$ and Ye Yuan$^{2}$
\thanks{$^{1}$Yi Lan and $^{2}$Ye Yuan are with the College of Computer and Information Science, Southwest University, Tiansheng Road, Chongqing, 400715, China
        {\tt\small yuanyekl@swu.edu.cn}}%
}

\usepackage{amsmath}   
\usepackage{amssymb}  
\usepackage{multirow}   
\usepackage{multicol}   
\usepackage{booktabs}   
\usepackage{array}      
\usepackage{tabularx}   
\usepackage{graphicx}
\usepackage{subcaption}
\usepackage[
backend=biber,      
style=ieee,         
sorting=none,
dashed=false         
]{biblatex}
\AtBeginBibliography{\footnotesize}

\begin{document}

\maketitle
\thispagestyle{empty}
\pagestyle{empty}

\begin{abstract}

Graph Neural Networks (GNNs) have achieved remarkable performance in graph representation learning, yet their inherent vulnerability to adversarial attacks poses severe security risks. Especially, black-box node injection attacks have become a major threat to GNNs since they inject malicious nodes without altering the original graph topology. However, they typically decouple the generation of malicious node features and edge connections, thereby resulting in suboptimal attack efficacy under stringent budgets. To address this critical issue, this study proposes a novel Target-aware Interaction-guided Reinforcement learning for Black-box node injection Attacks on GNNs (TIRBA), which formulates the attack as a Markov Decision Process and jointly optimizes node feature generation and edge construction in a heterogeneous action space. Firstly, TIRBA designs a target-aware interaction encoder to fuse information of node features and edges. Further, it introduces a class-center guidance mechanism to utilize prior class distribution information, thereby guiding efficient exploration of the high-dimensional feature space. Finally, a topology difference-aware state value evaluation is adopted to explicitly capture local structural anomalies caused by injected nodes, thereby stabilizing the reinforcement learning training process. Experimental results demonstrate that the proposed TIRBA significantly outperforms state-of-the-art black-box node injection attack methods.

\end{abstract}

\section{INTRODUCTION}

Graph Neural Networks (GNNs) have achieved strong performance on graph-structured data but remain vulnerable to adversarial attacks~\cite{zugnerAdversarialAttacksNeural, zhangTrustworthyGraphNeural2024}, highlighting the necessity of investigating graph representation vulnerabilities under diverse adversarial setups~\cite{G11, G12, G13, G14}. This inherent fragility has recently sparked comprehensive studies into certified robustness, privacy leakage, and generalized evaluation frameworks across various graph topologies~\cite{A5, A7, A10, A15}. Small perturbations may substantially degrade model predictions, raising concerns in security-sensitive applications such as fraud detection~\cite{G1, G2, G3}, as well as in other critical domains where adversarial instances manipulate social engagements, target model explanations, or disrupt dynamic link predictions~\cite{A6, A8, A9, A11}. Graph Modification Attacks (GMA) alter existing edges or node attributes~\cite{sharmaTaskModelAgnostic2023, zhuSimpleEfficientPartial2024}, and therefore require modification permissions that may be unavailable in practical adversarial scenarios~\cite{G15, G16, G17, G18}. By contrast, Node Injection Attacks (NIA) add fabricated nodes and incident edges while preserving all pre-existing nodes, attributes, and edges~\cite{zhaoHighlyImperceptibleBlackbox2025}. For example, an adversary may create new accounts and connect them to existing users to influence a graph-based decision system~\cite{G59, G60}. NIA is therefore feasible when adversaries can create new entities but cannot modify historical graph records~\cite{sunAdversarialAttackDefense2022}.

Despite its practicality, executing effective black-box NIA remains technically challenging due to the massive discrete search space and complex graph topologies~\cite{G19, G20, G21, G22}. First, attackers cannot access the victim model's architecture, parameters, gradients, or intermediate representations and must optimize the attack through prediction-score queries~\cite{sun2020adversarial}, a process that often leverages cluster priors, node voting mechanisms, or unnoticeable homogeneous node injections to circumvent defensive barriers~\cite{A3, A4, A13, A14}. Second, because GNNs couple structural and attribute information during message passing, node features and edges are deeply intertwined~\cite{G4, G5, G6}. However, existing methods often isolate feature generation and edge construction into separate stages, weakening their intrinsic coordination under tight query and perturbation budgets~\cite{sunAdversarialAttackDefense2022}. Furthermore, surrogate-based methods may exhibit limited transferability because of mismatched decision boundaries~\cite{zhangTrustworthyGraphNeural2024}, making direct black-box optimization highly desirable yet complex~\cite{G23, G24, G25, G26}. Searching a massive, discrete, and high-dimensional combinatorial space using only query feedback and without heuristic guidance also makes optimization inefficient and unstable~\cite{G7, G8}.

To address these issues, we propose TIRBA. TIRBA retains the standard Advantage Actor-Critic (A2C) objective~\cite{mnihAsynchronousMethodsDeep} but redesigns its actor and critic for feature generation and additional-edge selection in node injection. To coordinate heterogeneous generation, TIRBA introduces a target-aware interaction encoder that fuses parallel feature and topology representations, which helps overcome the information isolation problem in sequential generation~\cite{G27, G28, G29, G30}. Additionally, a Class-Center Guidance Mechanism utilizes prior class vectors to steer feature generation and applies Gumbel-Softmax for end-to-end differentiability. Finally, a Topology Difference-Aware State Value Evaluation uses the representation discrepancy between the target and its original neighbors as an explicit signal for state-value estimation. The main contributions of this paper are summarized as follows:
\begin{itemize}
\item We propose TIRBA, a black-box node injection attack framework based on the A2C algorithm~\cite{mnihAsynchronousMethodsDeep}. It formulates feature generation and edge selection as a Markov Decision Process and performs end-to-end optimization in a heterogeneous action space.
\item We introduce a target-aware interaction encoder, label-informed class-center guidance, and topology difference-aware state value evaluation. These mechanisms adapt the actor and critic to feature-topology dependencies in node injection.
\item We conduct extensive experiments on four real-world benchmark datasets with several widely used GNN models. The results demonstrate that TIRBA achieves superior attack performance compared with existing state-of-the-art methods.
\end{itemize}

\section{RELATED WORK}

\subsection{Adversarial Attacks}

With the widespread adoption of deep learning on non-Euclidean data, the robustness of GNNs has received extensive attention~\cite{G31, G32, G33, G34}. Existing adversarial attacks on graphs are broadly categorized into evasion attacks and poisoning attacks~\cite{sunAdversarialAttackDefense2022}. Early studies predominantly focused on GMA~\cite{sharmaTaskModelAgnostic2023},~\cite{zhuSimpleEfficientPartial2024}. For instance, the projected gradient descent (PGD)~\cite{chenUnderstandingImprovingGraph2022, xuTopologyAttackDefense2019} method leverages the gradient information of the target model to guide the attack. It modifies original edges or node features in the graph to mislead the model's predictions~\cite{G35, G36, G37, G38}. Such gradient-based attacks can be white-box when differentiating through the victim model or transfer-based black-box when using a surrogate, while GMA still requires permission to modify pre-existing graph elements~\cite{zugnerAdversarialAttacksNeural, sunAdversarialAttackDefense2022}. In actual security-sensitive scenarios, such global modifications can easily trigger alarms in defense systems~\cite{G39, G40, G41, G42}. Consequently, the practical applicability of these attacks is profoundly limited~\cite{zhangTrustworthyGraphNeural2024}.

\subsection{Node Injection Attacks}
To overcome the practical limitations of GMA, NIA was introduced~\cite{zhaoHighlyImperceptibleBlackbox2025}. In NIA, the attacker is only permitted to inject a limited number of fabricated nodes into the graph and cannot alter any original benign structures~\cite{G43, G44, G45, G46}. This paradigm renders the attack stealthier~\cite{sunAdversarialAttackDefense2022}. In recent years, several NIA methods have been proposed to evaluate and expose GNN security risks~\cite{zhaoHighlyImperceptibleBlackbox2025, G47, G48, G49, G50}. For example, TDGIA employs a heuristic edge connection rule based on topological defects~\cite{zouTDGIAEffectiveInjection2021}. GCIA incorporates graph contrastive learning to enhance query efficiency in a black-box setting~\cite{liuGCIABlackboxGraph}. G$^2$A2C adopts an Actor-Critic reinforcement learning architecture, formulating the attack as a sequential decision problem~\cite{juLetGraphBe2023}.Notably, reinforcement learning has demonstrated immense potential in orchestrating versatile black-box and universal graph attacks, while also prompting cutting-edge research into its own specific adversarial and backdoor vulnerabilities~\cite{A1, A2, A12, A16}. 

Sequential node injection methods commonly generate features before performing the subsequent edge-wiring decisions, which may weaken coordination between the two stages. To mitigate this problem, TIRBA introduces a target-aware interaction encoder to exchange information between the feature and topology branches. It further uses class-center guidance and topology difference-aware state value evaluation. Each injected node is first connected to the target by a mandatory anchoring edge, while the learned edge policy selects additional endpoints only when the edge budget exceeds one. The actor and critic are optimized jointly through score-based rewards without differentiating through the victim model.

\section{PRELIMINARY}

\subsection{Graph Neural Networks} 

We define an undirected attributed graph $G = (A, X)$, where $A \in \{0,1\}^{N \times N}$ is the adjacency matrix corresponding to the edge set $E$, and $X \in \mathbb{R}^{N \times d}$ is the node feature matrix. Specifically, $A_{ij} = 1$ if an edge exists, and $A_{ij} = 0$ otherwise.
GNN learns node representations by aggregating information from local neighbors. For a GNN with $L$ layers, the message passing process of the feature representation $H^{(l)}$ at layer $l$ is defined as:

\begin{equation}
H^{(l)} =
\sigma \left(
\text{AGGREGATE}\left(A, H^{(l-1)}\right) W^{(l)}
\right)
\end{equation}
where $H^{(0)} = X$ denotes the initial feature matrix, $\sigma(\cdot)$ is the activation function, $W^{(l)}$ is the weight matrix, and $\text{AGGREGATE}(\cdot)$ signifies the aggregation function.
For example, in a Graph Convolutional Network~\cite{kipfSemisupervisedClassificationGraph2017}, the propagation rule is given by:

\begin{equation}
H^{(l)} =
\sigma \left(
\tilde{D}^{-\frac{1}{2}}
\tilde{A}
\tilde{D}^{-\frac{1}{2}}
H^{(l-1)}
W^{(l)}
\right)
\end{equation}
where $\tilde{A} = A + I_N$, and $\tilde{D}$ is the corresponding degree matrix of $\tilde{A}$. In this paper, the experiments evaluate several models, such as GCN~\cite{kipfSemisupervisedClassificationGraph2017}, SGC~\cite{wuSimplifyingGraphConvolutional}, APPNP~\cite{gasteigerPredictThenPropagate2022}, and GAT~\cite{velickovicGraphAttentionNetworks2018}. All these models adhere to this general framework. For classification tasks, the model's output is mapped by the Softmax function to a probability distribution $Z \in \mathbb{R}^{N \times C}$, where $C$ is the number of classes. The fully trained black-box victim model is denoted as $f_{\theta^*}(\cdot)$.

\subsection{Attack Formulation}

This paper focuses on the Black-box Evasion Attack within a node classification scenario~\cite{yuGZOOBlackboxNode2025}. In this setting, the attacker lacks access to the internal structure, weight gradients, and specific training data of the victim model $f_{\theta^*}$, relying solely on external feedback~\cite{G51, G52, G53, G54}. They can only query the target model a limited number of times and utilize the output feedback to guide the attack strategy. To conduct the attack without perturbing original benign nodes, NIA permits the attacker to inject fabricated, malicious nodes into the original graph $G$~\cite{G55, G56, G57, G58}. Assuming the target node is $v_t$, the attacker's objective is to construct a feature matrix $X_a \in \mathbb{R}^{\mathcal{B}_n \times d}$ for the injected nodes and a connection matrix $A_a \in \{0,1\}^{\mathcal{B}_n \times N}$ linking the injected nodes to the original graph nodes. In this work, we focus on a single-injected-node setting ($\mathcal{B}_n = 1$). This yields a newly modified graph $G' = (A', X')$:

\begin{equation}
A' = \begin{bmatrix} A & A_a^T \\ A_a & 0 \end{bmatrix}, \quad X' = \begin{bmatrix} X \\ X_a \end{bmatrix}
\label{eq:injected_graph_matrix}
\end{equation}
To maintain stealthiness, injected nodes remain disconnected from each other. The injection is subjected to strict budgets: the node budget is $\mathcal{B}_n$, the edge budget is $\mathcal{B}_e$, and the feature budget is $\mathcal{B}_x$. Here, $\mathcal{B}_e$ denotes the maximum number of edges incident to each injected node, including the mandatory target-anchoring edge. Specifically, the number of active feature dimensions in the injected feature matrix $X_a$ is constrained by $\sum_{i=1}^{\mathcal{B}_n} \sum_{j=1}^{d} \mathbb{I}(X_a^{(i,j)} > 0) \le \mathcal{B}_x$. The attack aims to identify the optimal $X_a$ and $A_a$ within these budgets to maximize the prediction loss on $v_t$ and induce misclassification. Given the vast discrete search space, we formulate this challenge as an MDP.

\section{METHODOLOGY}

\subsection{Attack Framework}

\begin{figure*}[tp] 
    \centering
    \includegraphics[width=0.9\textwidth]{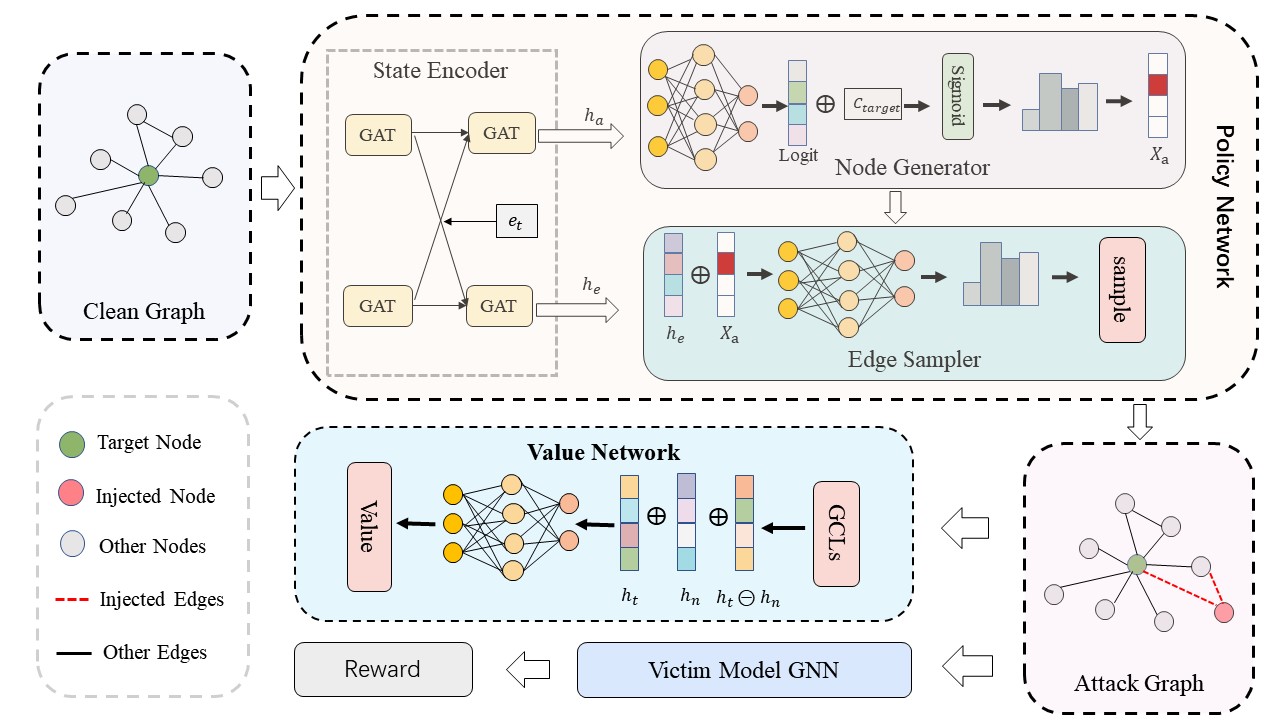} 
    
    \caption{Overview of the TIRBA Framework for Black-Box Node Injection Attacks.}
    
    \label{fig:framework}
\end{figure*}

Since generating discrete features and edges in a black-box setting is inherently a sequential decision-making process, we formulate it as an MDP and propose TIRBA, an end-to-end Actor-Critic framework (Fig. \ref{fig:framework}). The Policy Network (Actor) uses a class center vector $C_{target}$ to generate injected node features and edges. In the single-injected-node setting, the generated feature vector is denoted by $X_a \in \mathbb{R}^{d}$. The resulting Attack Graph is evaluated via two feedback loops: querying the victim GNN yields the loss variation as the reward, while the Value Network (Critic) calculates the local topology difference $h_t \ominus h_n$ to estimate the state value. This process iterates until target misclassification or budget exhaustion.

We mathematically define the node injection attack on the victim GNN through the following three core components:

\textbf{State.} 
The state $s_t$ at time step $t$ corresponds to a local sub-graph centered on the target node $v_t$. This state encapsulates both the local structural matrix and the node feature matrix, providing the agent with a comprehensive observation of the environment.

\textbf{Action.} 
The action sequence first samples a relaxed feature vector $X_a$ for the injected node. The injected node is then linked to $v_t$ by a mandatory target-anchoring edge. If $\mathcal{B}_e>1$, each remaining edge action selects an unconnected original node $v_c$ from the candidate set $\mathcal{V}_c$.

\textbf{Reward.} 
As an untargeted evasion attack, the reward relies on the Cross-Entropy Loss $\mathcal{L}$ variation regarding the victim's initial prediction for $v_t$, strictly avoiding ground-truth labels. The immediate reward $r_t$ is the loss discrepancy:
\begin{equation}
r_t = \mathcal{L}(v_t, G'_{t+1}) - \mathcal{L}(v_t, G'_t)
\label{eq:reward}
\end{equation}
where $G'_t$ and $G'_{t+1}$ denote the graph states before and after executing the action $a_t$.

\subsection{Policy Network}

The Policy Network aims to learn the optimal injection strategy. It orchestrates the generation of the injected node's attributes and topology through three primary modules:

\textbf{State Encoder.}
To achieve enhanced target focus, we adopt target-conditioned message passing. Two parallel GATs extract the initial feature embedding $h_a^{(1)}$ and edge embedding $h_e^{(1)}$. To facilitate context sharing across both tasks, we concatenate them at the intermediate layer. Subsequently, we explicitly inject the target node's initial projection embedding $e_t$ (projected to match the concatenated dimension) to realize target-aware interaction fusion:
\begin{equation}
\tilde{h}_a = (h_a^{(1)} \parallel h_e^{(1)}) + e_t
\end{equation}
\begin{equation}
\tilde{h}_e = (h_e^{(1)} \parallel h_a^{(1)}) + e_t
\end{equation}
where $\parallel$ denotes the concatenation operation. The fused representations are then aggregated by the second GAT layer, which outputs the final state encoding $h_a$ for feature generation and $h_e$ for edge sampling.

\textbf{Node Generator.}
Exploring a high-dimensional discrete feature space in a black-box setting is highly inefficient. Therefore, we introduce a heuristic guidance mechanism. To cross the decision boundary with minimal cost, we adopt a pseudo-targeted strategy by selecting the second-most probable class as the target class. This pseudo-targeted design acts as an efficient heuristic to rapidly cross the decision boundary for untargeted evasion goals. The decoder first projects $h_a$ to an initial logit vector $z$. We then explicitly incorporate the feature center vector $C_{target}$ of this target class to obtain a guided distribution $z_g$. To maintain black-box strictness, $C_{target}$ is estimated by averaging features of accessible nodes assigned to this pseudo-class:
\begin{equation}
z_g = z + \alpha \cdot C_{target}
\end{equation}
where $\alpha$ is a weight parameter controlling the guidance intensity. We apply a continuous relaxation to enable differentiable multi-hot feature sampling. The final relaxed feature row vector $X_a$ of the injected node is derived as:
\begin{equation}
X_a = \text{Sigmoid}\left(\frac{z_g + g}{\tau}\right)
\end{equation}
where $g$ represents independent noise sampled from the standard Gumbel distribution, and $\tau$ is a temperature hyperparameter controlling distribution smoothness.

\textbf{Edge Sampler.}
When $\mathcal{B}_e>1$, an MLP computes the additional-edge score $s_c$ for each candidate original node $v_c \in \mathcal{V}_c$. The scorer combines the topology representation $h_e$ and the newly generated feature $X_a$:
\begin{equation}
s_c = \text{MLP}_e \left( h_e^{(v_t)} \parallel h_e^{(v_c)} \parallel (X_a W_p) \right)
\end{equation}
where $h_e^{(v_t)}$ and $h_e^{(v_c)}$ represent the specific state encodings of the target node and candidate node, respectively, and $W_p$ is a feature projection matrix. The resulting scores are transformed into an action probability distribution via Softmax:
\begin{equation}
P(v_c | s_t) = \frac{\exp(s_c)}{\sum_{k \in \mathcal{V}_c} \exp(s_k)}
\end{equation}

\subsection{Value Network}

The Value Network estimates the expected return to supervise the Policy Network. Since conventional mean aggregation over-smooths representations and obscures injected structural anomalies, we refine the pooling strategy. We explicitly compute the difference $h_t \ominus h_n$ between the target node's hidden state $h_t$ and its benign neighbors' mean state $h_n$. This directly captures the broken homophily caused by injected edges~\cite{G9, G10}. Finally, the state value $V_\phi(s_t)$ is evaluated by concatenating these components into an MLP:
\begin{equation}
V_\phi(s_t) = \text{MLP}_v \left( h_t \parallel h_n \parallel (h_t \ominus h_n) \right)
\end{equation}
where $\ominus$ denotes element-wise subtraction, and $\phi$ represents the learnable parameters of the Critic network. By explicitly capturing this topological discrepancy, the Value Network directly perceives local structural variations caused by the injected node, facilitating faster convergence of the reinforcement learning process.

\subsection{Training Algorithm}

TIRBA is trained end-to-end utilizing the Advantage Actor-Critic algorithm~\cite{mnihAsynchronousMethodsDeep}. Upon the completion of each attack trajectory, we first compute the cumulative reward $R_t$ with a discount factor $\gamma$:

\begin{equation}
R_t = \sum_{k=0}^{\infty} \gamma^k r_{t+k}
\label{eq:return}
\end{equation}
where $r_{t+k}$ represents the immediate reward at future step $k$. Subsequently, we leverage the baseline from the Critic network to derive the advantage function $\mathcal{A}(s_t, a_t)$:

\begin{equation}
\mathcal{A}(s_t, a_t) = R_t - V_\phi(s_t)
\label{eq:advantage}
\end{equation}

The policy loss aims to maximize the expected reward:

\begin{equation}
\mathcal{L}_p =
-\frac{1}{T}
\sum_{t=1}^{T}
\log \pi_\theta(a_t | s_t) \mathcal{A}(s_t, a_t)
\label{eq:policy_loss}
\end{equation}

The value loss optimizes the Critic to accurately estimate the state value:

\begin{equation}
\mathcal{L}_v =
\frac{1}{T}
\sum_{t=1}^{T}
\text{SmoothL1}\left(V_\phi(s_t) - R_t\right)
\label{eq:value_loss}
\end{equation}

To enforce the feature budget constraint and preserve attack stealthiness, we introduce a soft penalty formulation regulating the generated values:

\begin{equation}
\mathcal{L}_f =
\max \left( 0, \sum_{i=1}^{\mathcal{B}_n} \sum_{j=1}^{d} X_a^{(i,j)} - \mathcal{B}_x \right)
\label{eq:feature_budget}
\end{equation}
where $X_a^{(i,j)}$ is the value of the $j$-th dimension in the $i$-th generated feature vector, and $d$ is the total feature dimensionality. $\mathcal{B}_x$ denotes the node feature perturbation budget predefined by the victim system.

The total optimization objective of the model, denoted as $\mathcal{L}_{total}$, comprises the policy loss, value loss, and the feature budget penalty:

\begin{equation}
\mathcal{L}_{total} =
\mathcal{L}_p + \lambda_v \mathcal{L}_v + \lambda_f \mathcal{L}_f
\label{eq:total_loss}
\end{equation}
where $\lambda_v$ and $\lambda_f$ are hyperparameters balancing the respective loss components.

\section{EXPERIMENTS}
In this section, we comprehensively evaluate TIRBA on four benchmark datasets and compare it against existing state-of-the-art methods. Our experiments are designed to address the following four key research questions: \textbf{(RQ1)} Under the strict black-box single-node injection setting, can TIRBA achieve superior attack performance against various GNN victim models? \textbf{(RQ2)} Do the core components of the TIRBA framework (e.g., target-aware mechanism, guidance mechanism, and topology difference evaluation) effectively contribute to its overall attack efficacy? \textbf{(RQ3)} How does TIRBA's attack performance fluctuate under varying budget constraints (such as node, edge, and feature budgets)? \textbf{(RQ4)} In practical scenarios, how does TIRBA visually alter the target node's feature distribution to induce misclassification?

\subsection{Experimental Setup}

\textbf{Datasets.}
We evaluated the performance of TIRBA on four widely-used real-world graph datasets. These datasets constitute citation networks: Cora, Citeseer~\cite{senCollectiveClassificationNetwork2008}, Cora-ML~\cite{mccallumAutomatingConstructionInternet}, and Pubmed~\cite{senCollectiveClassificationNetwork2008}. Detailed dataset statistics are summarized in Table \ref{table_datasets}.

\begin{table}[h]
\caption{Statistics of the Datasets}
\label{table_datasets}
\begin{center}
\begin{tabular}{l|cccc}
\hline
& \textbf{Nodes} & \textbf{Edges} & \textbf{Features} & \textbf{Classes}\\
\hline
\textbf{Cora} & 2708 & 5429 & 1433 & 7\\
\textbf{Citeseer} & 3327 & 4732 & 3703 & 6\\
\textbf{Cora-ML} & 2995 & 8416 & 2879 & 7\\
\textbf{Pubmed} & 19717 & 44338 & 500 & 3\\
\hline
\end{tabular}
\end{center}
\end{table}

\textbf{Victim Models.}
To verify the generalization capability of the proposed attack, we selected four mainstream Graph Neural Network architectures as victim models: GCN~\cite{kipfSemisupervisedClassificationGraph2017}, SGC~\cite{wuSimplifyingGraphConvolutional}, APPNP~\cite{gasteigerPredictThenPropagate2022}, and GAT~\cite{velickovicGraphAttentionNetworks2018}.

\textbf{Baselines.}
To evaluate TIRBA, we employ Clean (no attack) and Random baselines alongside four advanced node injection attacks: heuristic-based TDGIA~\cite{zouTDGIAEffectiveInjection2021}, gradient-optimized PGD~\cite{chenUnderstandingImprovingGraph2022, xuTopologyAttackDefense2019}, contrastive-driven GCIA~\cite{liuGCIABlackboxGraph}, and RL-based G$^2$A2C~\cite{juLetGraphBe2023}. All experiments strictly constrain the attack budget per target to 1 injected node and 1 edge.

\subsection{Performance Comparison}

To comprehensively validate TIRBA's attack capabilities under a strict black-box single-node injection setting, we conducted extensive evaluations against five diverse baseline methods. The experiments span four widely adopted benchmark datasets (Cora, Citeseer~\cite{senCollectiveClassificationNetwork2008}, Cora-ML, Pubmed~\cite{senCollectiveClassificationNetwork2008}) and four representative GNN architectures (GCN~\cite{kipfSemisupervisedClassificationGraph2017}, SGC~\cite{wuSimplifyingGraphConvolutional}, APPNP~\cite{gasteigerPredictThenPropagate2022}, GAT~\cite{velickovicGraphAttentionNetworks2018}). As quantitatively detailed in Table \ref{table_mr}, TIRBA consistently achieves the highest Misclassification Rate across various dataset and model combinations, demonstrating an exceptional generalization capability. Particularly in environments with challenging discrete feature spaces, TIRBA exhibits superior stability and adversarial strength. For instance, it markedly outperforms the highly competitive second-best baseline, PGD~\cite{chenUnderstandingImprovingGraph2022, xuTopologyAttackDefense2019}, on the Cora dataset when attacking the SGC model. Furthermore, its attack success rate approaches an impressive 94\% on the Pubmed~\cite{senCollectiveClassificationNetwork2008} dataset. These compelling results conclusively demonstrate TIRBA's robust efficacy and adaptability, even when operating under stringent query constraints and extreme injection budgets. 
In response to \textbf{RQ1}: In highly restrictive single-node black-box scenarios, TIRBA consistently and significantly outperforms all evaluated baselines. This explicitly proves its outstanding capability in jointly generating robust adversarial node features and edges under tight limitations.

\begin{table}[h]
\centering
\caption{Comparison of Misclassification Rate (\%) of different injection attack methods on benchmark datasets and target models. Bold numbers show the best performance.}
\label{table_mr}
\footnotesize
    \begin{tabular}{clcccc}
    \toprule
    \textbf{Victim Model} & \textbf{Attacks} & \textbf{Cora} & \textbf{Citeseer} & \textbf{Cora-ML} & \textbf{Pubmed} \\ 
    \midrule
    & Clean & 19.80 & 31.60 & 21.50 & 23.70 \\
    & Random & 23.66 & 37.00 & 22.83 & 62.80 \\
    & GCIA & 34.52 & 45.60 & 28.26 & 77.86 \\
    \textbf{GCN} & PGD & 47.97 & 58.60 & 52.38 & 69.22 \\
    & TDGIA & 28.60 & 41.22 & 25.84 & 70.00 \\
    & G$^2$A2C & 33.10 & 53.70 & 35.21 & 81.24 \\
    & \textbf{Ours} & \textbf{63.80} & \textbf{73.98} & \textbf{58.93} & \textbf{88.76} \\
    \midrule
    & Clean & 21.50 & 32.90 & 24.29 & 22.90 \\
    & Random & 26.14 & 38.10 & 27.26 & 63.64 \\
    & GCIA & 37.44 & 45.36 & 37.60 & 76.94 \\
    \textbf{SGC} & PGD & 48.28 & 59.53 & 55.04 & 66.30 \\
    & TDGIA & 30.44 & 42.14 & 28.74 & 70.18 \\
    & G$^2$A2C & 38.56 & 53.64 & 36.26 & 71.90 \\
    & \textbf{Ours} & \textbf{69.06} & \textbf{74.04} & \textbf{59.96} & \textbf{81.24} \\
    \midrule
    & Clean & 31.40 & 36.70 & 24.80 & 28.10 \\
    & Random & 34.36 & 40.32 & 27.52 & 61.04 \\
    & GCIA & 38.72 & 44.52 & 31.34 & 65.94 \\
    \textbf{APPNP} & PGD & 40.21 & 54.47 & 49.58 & 69.30 \\
    & TDGIA & 42.44 & 47.24 & 32.27 & 68.18 \\
    & G$^2$A2C & 34.14 & 51.36 & 36.74 & 83.16 \\
    & \textbf{Ours} & \textbf{68.72} & \textbf{68.26} & \textbf{54.88} & \textbf{86.84} \\
    \midrule
    & Clean & 21.00 & 31.30 & 18.64 & 23.60 \\
    & Random & 22.02 & 35.02 & 20.14 & 63.42 \\
    & GCIA & 27.64 & 41.24 & 25.11 & 76.86 \\
    \textbf{GAT} & PGD & 44.76 & 51.18 & 42.46 & 75.00 \\
    & TDGIA & 28.78 & 39.28 & 23.41 & 79.92 \\
    & G$^2$A2C & 26.54 & 55.52 & 43.05 & 87.84 \\
    & \textbf{Ours} & \textbf{51.08} & \textbf{64.14} & \textbf{46.99} & \textbf{93.98} \\
    \bottomrule
    \end{tabular}
\end{table}

\subsection{Ablation Study}

\begin{figure}[tp] 

    \centering

    \includegraphics[width=0.6\textwidth]{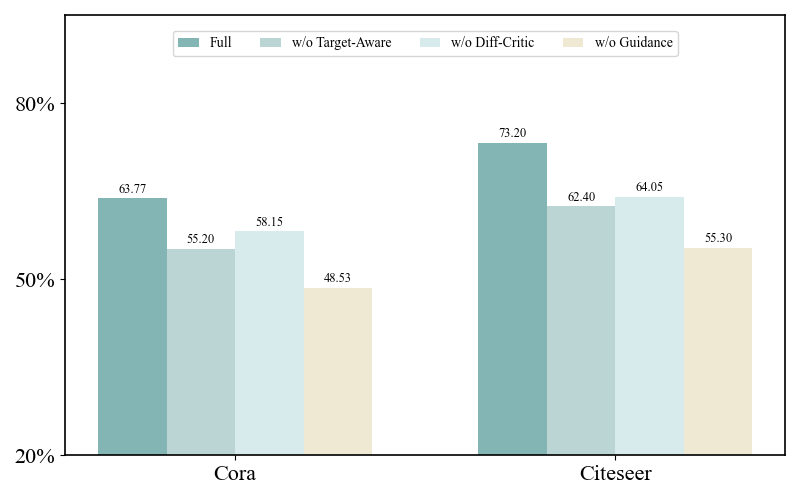} 

    \caption{Ablation study of the TIRBA framework on Cora and Citeseer datasets.}

    \label{fig:ablation}

\end{figure}
We conducted an ablation study on the Cora and Citeseer datasets~\cite{senCollectiveClassificationNetwork2008} to validate TIRBA's components (Fig. \ref{fig:ablation}). The full model consistently outperforms its variants. Removing the class-center guidance (w/o Guidance) causes the severest performance drop, highlighting its necessity in reducing the search space. Omitting the target-aware interaction encoder (w/o Target-Aware) significantly lowers the Misclassification Rate, proving that target focus is essential to break local homophily. Finally, excluding the difference-aware evaluation (w/o Diff-Critic) degrades performance, validating its ability to capture topological anomalies. In response to \textbf{RQ2}: All three core components are indispensable for TIRBA's robust attack efficacy.

\subsection{Budget Analysis}

\begin{figure}[tp] 
    \centering
    \begin{subfigure}[b]{0.4\textwidth}
        \centering
        \includegraphics[width=1.0\textwidth]{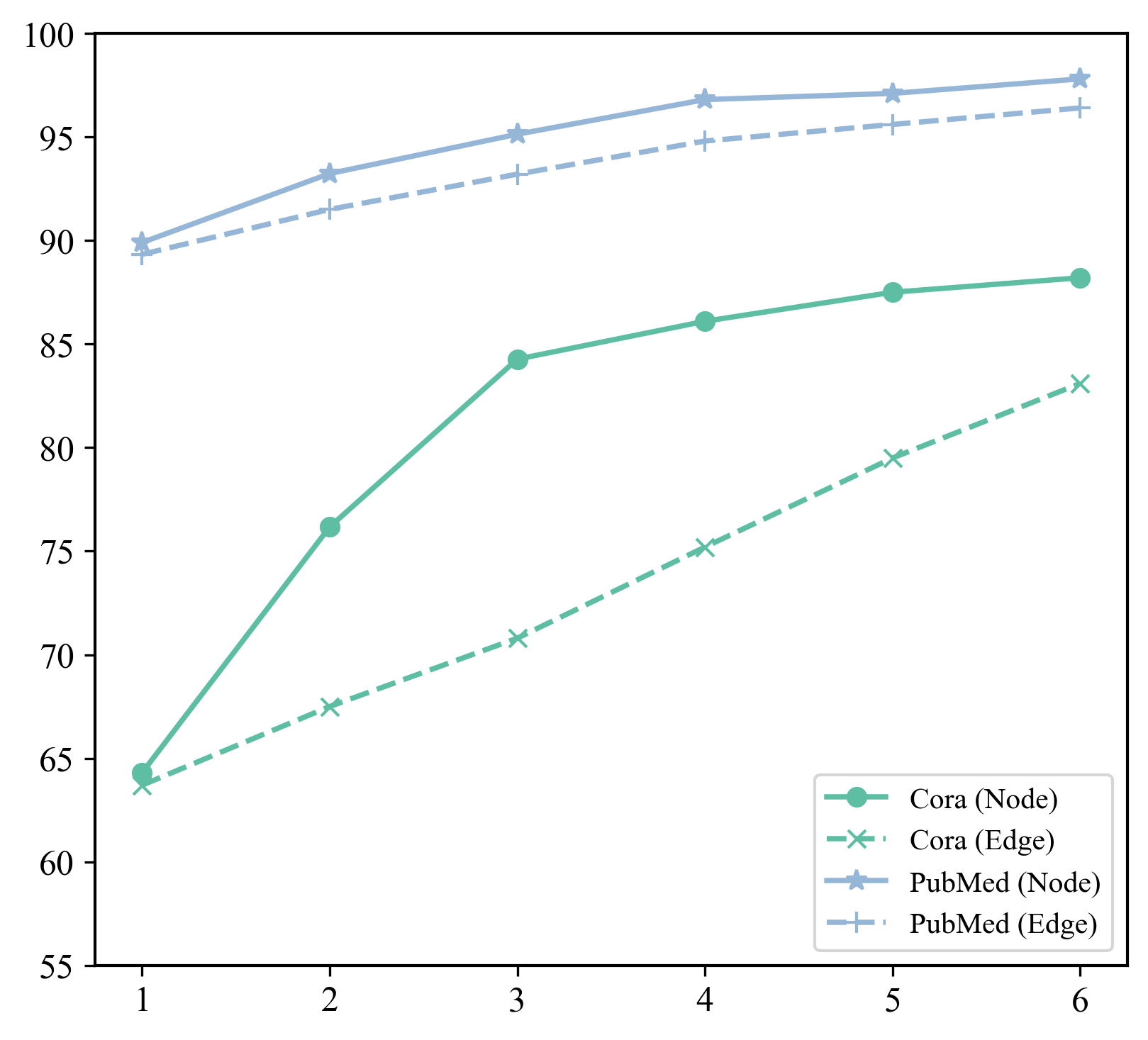}
        \caption{Node/Edge Budget}
        \label{fig:line_a}
    \end{subfigure}
    \hfill 
    \begin{subfigure}[b]{0.4\textwidth}
        \centering
        \includegraphics[width=1.0\textwidth]{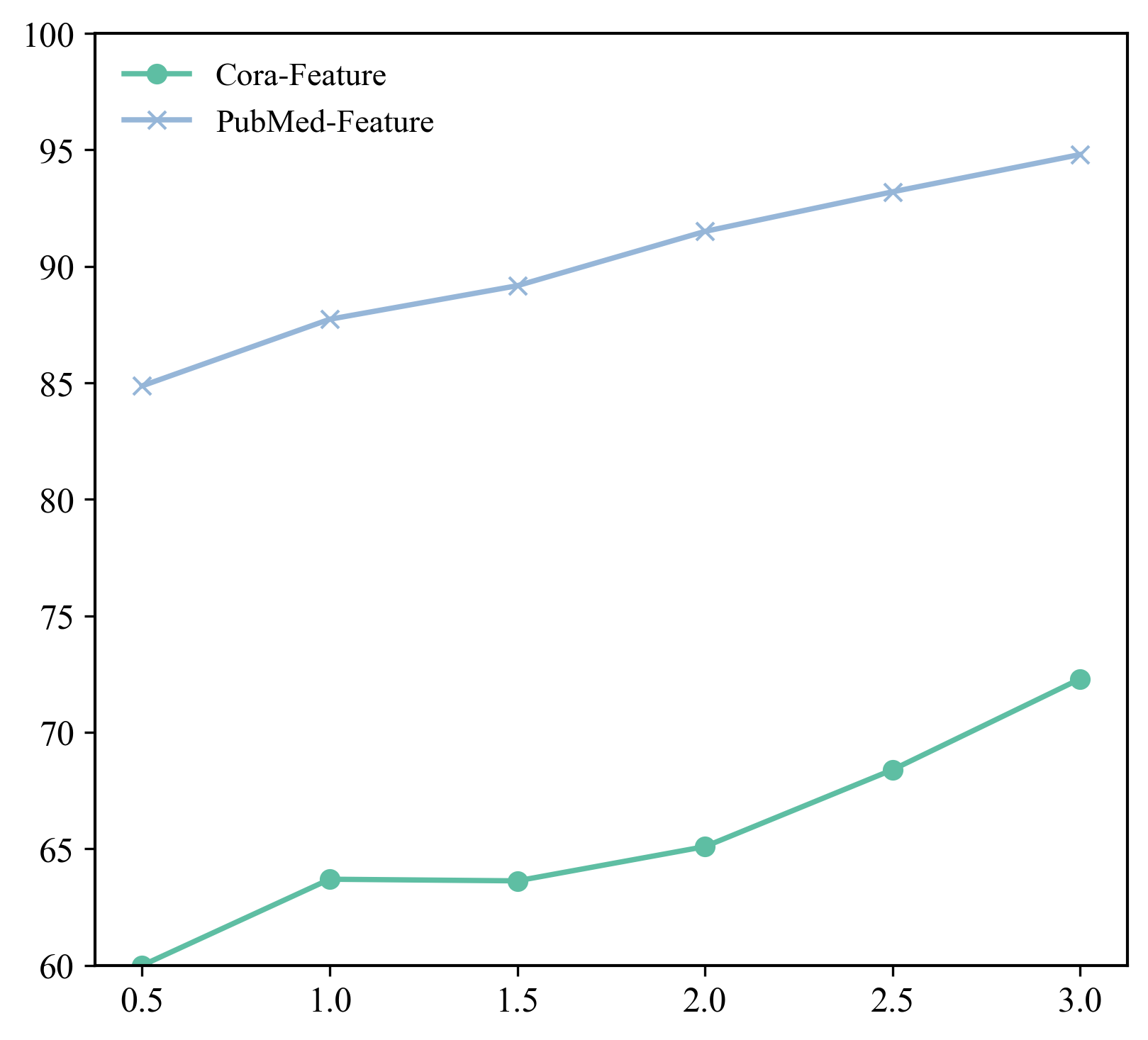}
        \caption{Feature Budget}
        \label{fig:line_b}
    \end{subfigure}
    
    \caption{The influence of different attack budgets on the Misclassification Rate. }
    \label{fig:line}
\end{figure}

To evaluate how attack budgets affect TIRBA's efficacy, we analyzed the Misclassification Rate on the Cora and PubMed~\cite{senCollectiveClassificationNetwork2008} datasets. As shown in Fig. \ref{fig:line}(\subref{fig:line_a}), increasing node or edge budgets independently causes the misclassification rate to rise steadily before plateauing, indicating diminishing returns. Similarly, Fig. \ref{fig:line}(\subref{fig:line_b}) illustrates that increasing the feature generation budget quickly saturates the attack performance. Overall, TIRBA maintains a high misclassification rate even under strict, low-budget conditions. 
In response to \textbf{RQ3}: TIRBA's attack power rises steadily and plateaus as node, edge, or feature budgets increase. Crucially, it achieves high attack success rates even under extremely tight budget limits.

\subsection{Case Study}
\begin{figure}[tp] 
    \centering
    \begin{subfigure}[b]{0.4\textwidth}
        \centering
        \includegraphics[width=1.0\textwidth]{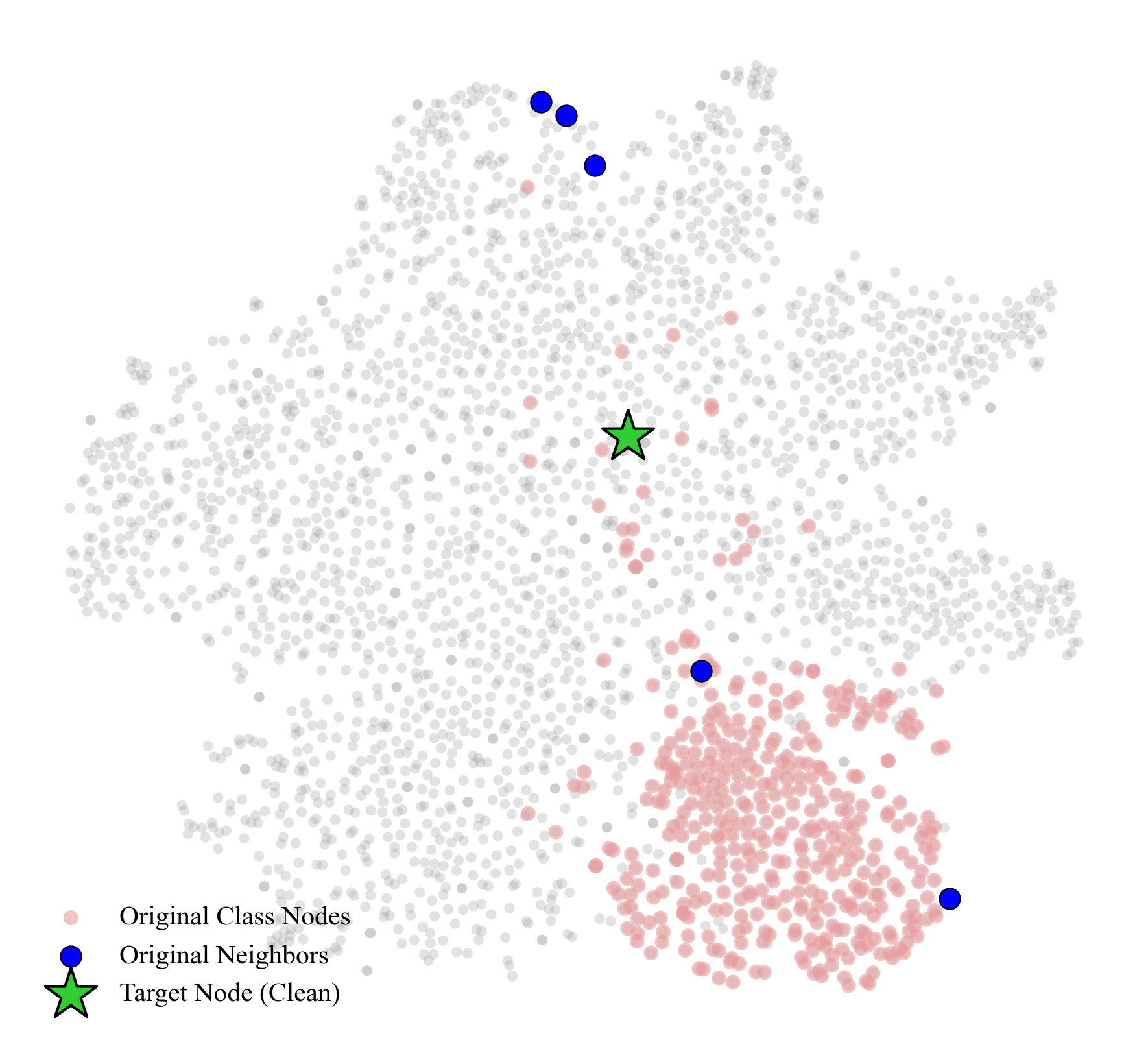}
        \caption{Before the attack}
        \label{fig:clean}
    \end{subfigure}
    \hfill 
    \begin{subfigure}[b]{0.4\textwidth}
        \centering
        \includegraphics[width=1.0\textwidth]{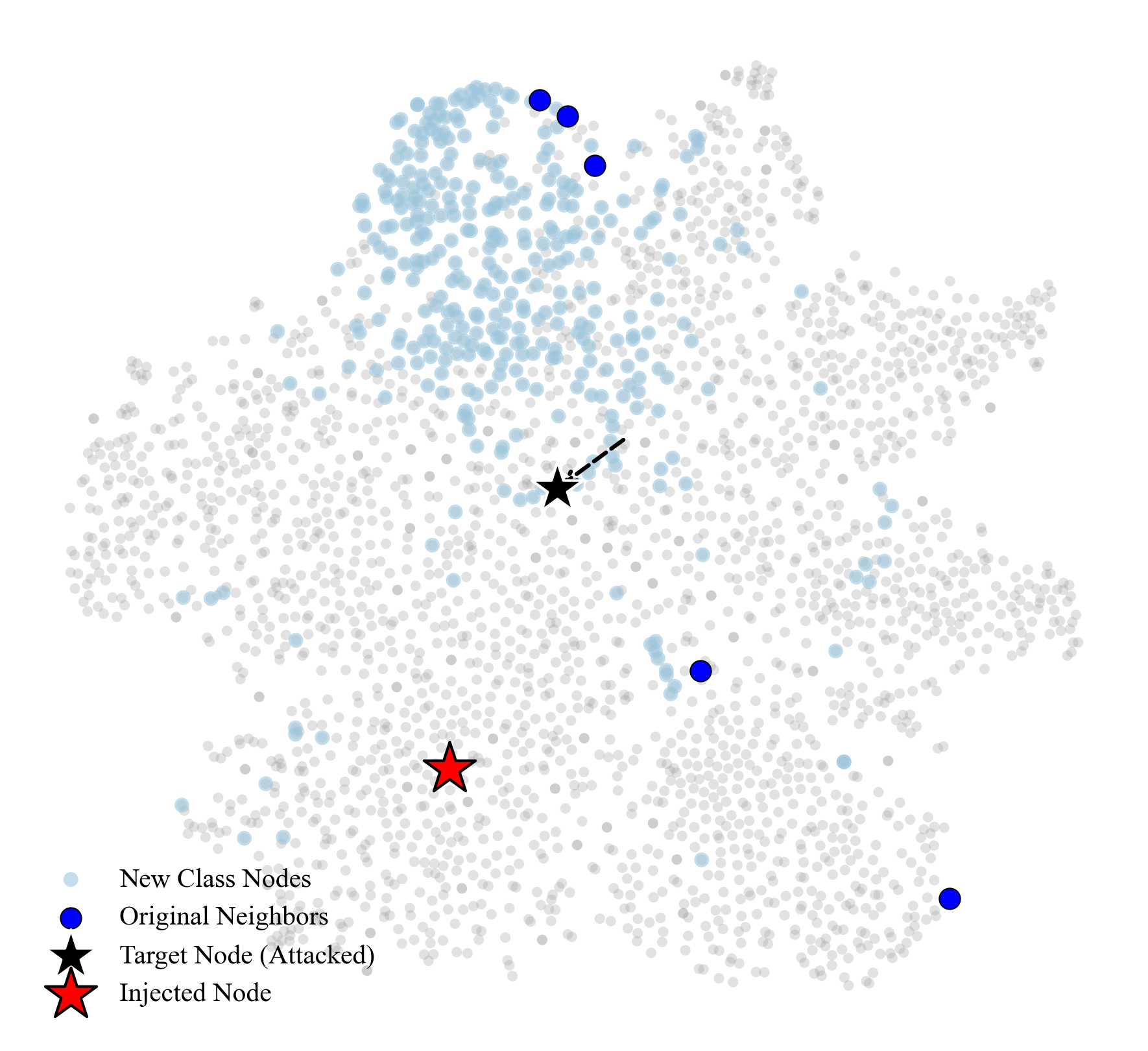}
        \caption{After the attack}
        \label{fig:injected}
    \end{subfigure}
    
    \caption{Visualization of the attack.}
    \label{fig:attack}
\end{figure}

To visualize TIRBA's attack effect, we employ T-SNE to map feature embeddings before and after node injection. 
Fig. \ref{fig:attack}(\subref{fig:clean}) visualizes the clean graph's feature space, where the original target node (green star) resides accurately within its true class cluster (light red dots), surrounded by its intrinsic neighbors (blue dots). 
Fig. \ref{fig:attack}(\subref{fig:injected}) displays the post-attack feature space. The TIRBA-injected malicious node (red star) strongly perturbs the target node (black star). Tracking the black dashed line, the target node distinctly shifts from its original cluster and merges into an incorrect class area (light blue dots). 
This noticeable displacement confirms that TIRBA effectively alters target representations via topological perturbation, successfully inverting the model's prediction.
In response to \textbf{RQ4}: Visual analysis verifies that TIRBA's injected nodes aggressively pull the target node across the feature space, forcing it to assimilate into a new class cluster and successfully inducing misclassification.

\section{CONCLUSIONS}

This paper introduces TIRBA, an end-to-end reinforcement learning framework for investigating GNN security vulnerabilities under black-box node injection attacks. Built on the standard A2C objective, TIRBA models feature generation and edge construction within a unified MDP and adapts the actor and critic to the heterogeneous graph attack space. To improve optimization in this large search space, the framework integrates three key components: a target-aware interaction encoder for feature-topology fusion, a class-center guidance mechanism for directed exploration, and a topology difference-aware critic for state-value estimation. Extensive experiments show that TIRBA consistently achieves higher misclassification rates than the evaluated baselines under restrictive attack budgets and across multiple victim architectures. It effectively induces prediction errors on target nodes while preserving the original graph elements. These results demonstrate the effectiveness, adaptability, and practical potential of TIRBA for analyzing and exposing the robustness limitations of GNN-based systems.Future work will explore stealthier and more effective NIA strategies.


\clearpage
\printbibliography
\end{document}